\documentclass{ecai} 

\usepackage{latexsym}
\usepackage{amssymb}
\usepackage{amsmath}
\usepackage{amsthm}
\usepackage{booktabs}
\usepackage{enumitem}
\usepackage{graphicx}
\usepackage{color}
\usepackage{booktabs,tabularx,array, makecell, multirow}

\newcommand{\BibTeX}{B\kern-.05em{\sc i\kern-.025em b}\kern-.08em\TeX}

\begin{document}


\begin{frontmatter}


\paperid{123} 


\title{A Multi-Objective Genetic Algorithm for Healthcare Workforce Scheduling}


\author[A]{\fnms{Vipul}~\snm{Patel}\thanks{Corresponding Author. Email: vipul.patel@quantiphi.com}\footnote{Accepted at the Multi-Objective Decision Making Workshop (MODeM 2025) at ECAI 2025.}}
\author[B]{\fnms{Anirudh}~\snm{Deodhar}}
\author[C]{\fnms{Dagnachew}~\snm{Birru}} 

\address[A]{Senior Machine Learning Engineer R\&D, Phi Labs, Quantiphi}
\address[B]{Principal Architect R\&D, Phi Labs, Quantiphi}
\address[C]{Global Head R\&D, Phi Labs, Quantiphi}



\begin{abstract}

Workforce scheduling in the healthcare sector is a significant operational challenge, characterized by fluctuating patient loads, diverse clinical skills, and the critical need to control labor costs while upholding high standards of patient care. This problem is inherently multi-objective, demanding a delicate balance between competing goals: minimizing payroll, ensuring adequate staffing for patient needs, and accommodating staff preferences to mitigate burnout. We propose a Multi-objective Genetic Algorithm (MOO-GA) that models the hospital unit workforce scheduling problem as a multi-objective optimization task. Our model incorporates real-world complexities, including hourly appointment-driven demand and the use of modular shifts for a multi-skilled workforce. By defining objective functions for cost, patient care coverage, and staff satisfaction, the GA navigates the vast search space to identify a set of high-quality, non-dominated solutions. Demonstrated on datasets representing a typical hospital unit, the results show that our MOO-GA generates robust and balanced schedules. On average, the schedules produced by our algorithm showed a 66\% performance improvement over a baseline that simulates a conventional, manual scheduling process. This approach effectively manages trade-offs between critical operational and staff-centric objectives, providing a practical decision support tool for nurse managers and hospital administrators.

\end{abstract}

\end{frontmatter}


\section{Introduction}

Effective workforce scheduling is paramount in the volatile healthcare sector, where daily and even hourly variations in patient volume and needs are standard. The quality of a schedule has direct consequences for patient outcomes, operational costs, and staff workload. However, constructing an optimal schedule is a notoriously difficult combinatorial problem. Managers must balance the skills and certifications of diverse employee groups— such as registered nurses, licensed practical nurses, and certified nursing assistants, along with managerial staff, against a backdrop of strict regulatory requirements and complex, modular shift structures.

The core difficulty lies in navigating the trade-offs between competing, often contradictory, goals. A focus on minimizing costs can lead to understaffing and jeopardize patient safety, while a schedule that prioritizes maximum coverage may become financially unviable. Furthermore, these operational objectives must be weighed against the personal preferences of the staff, whose satisfaction is key to mitigating high rates of burnout and turnover. This reality reframes the scheduling task as a classic multi-objective decision-making (MODeM) problem, requiring a delicate balance between:
\begin{itemize}
  \item Operational Efficiency: To control costs by aligning staffing precisely with demand.
  \item Quality of Patient Care: To meet or exceed standards for patient-to-staff ratios and required skill sets.
  \item Staff Satisfaction: To accommodate employee requests for shifts and leave, thereby improving morale and retention.
\end{itemize}

\subsection{Related Work}

The Workforce Scheduling Problem (WSP), particularly in healthcare contexts known as the Nurse Scheduling Problem (NSP), is a well-documented NP-hard combinatorial challenge that has served as a benchmark for numerous optimization methods. Broader solution approaches can be categorized into exact methods, metaheuristics, and artificial intelligence techniques \cite{ERNST2004}.

The challenge of workforce management is critical, especially in healthcare where high nurse burnout and turnover are prevalent. A 2023 McKinsey report\cite{McKinsey2023} indicates that manageable workloads and schedule flexibility are paramount for job satisfaction. This context underscores the urgent need for advanced scheduling optimization tools to address these industry-wide issues.

Exact methods, such as Mixed-Integer Linear Programming (MILP), are designed to find a mathematically proven optimal solution. However, their computational cost grows exponentially with problem size, making them impractical for the large-scale and dynamic scheduling required in most hospital units, which involve complex staff roles and regulatory constraints \cite{VANDENBERGH2013367}. Researchers have demonstrated the use of Linear Programming \cite{Kumar2014}, 0-1 Goal Programming to balance hospital objectives with nurse preferences \cite{AZAIEZ2005491}, and modern constraint-solving libraries like Google OR-Tools \cite{Oliveira2024}. Some have even extended the IP approach to incorporate novel concepts like "occupational justice" by basing shift preferences on nurse performance evaluations \cite{Pahlevanzadeh2021}. These studies highlight the foundational role of mathematical programming in this domain, but also underscore the need for more scalable methods for larger problems.

This limitation has led to the widespread adoption of heuristics and metaheuristics, which sacrifice a guarantee of absolute optimality for the ability to find high-quality solutions within a practical timeframe. Among these, Genetic Algorithms (GAs) have been particularly effective due to their robustness in handling the complex, non-linear constraints inherent in nurse rostering \cite{AICKELIN2004761}. Further research into improving GAs has focused on specialized operators and constraint-handling techniques, such as developing "repairing strategies" to convert infeasible solutions into feasible ones, a crucial mechanism for highly constrained problems \cite{Wasanapradit2011}. 

Some studies have also explored hybrid "matheuristic" approaches, combining exact models with heuristics to balance solution quality and computational speed \cite{TALARICO201565}. These methods combine different techniques to leverage their respective strengths. Strategies include two-stage heuristics that first generate a feasible schedule and then improve it with local search \cite{WONG201499} and cooperative methods that use IP to find strong lower bounds and CP to efficiently find feasible solutions \cite{RAHIMIAN201783}. 
A more recent and sophisticated trend involves leveraging machine learning \cite{Tianyu2021}, such as using Deep Neural Networks to intelligently guide heuristic selection and Mixed Integer Programming for diversification \cite{CHEN2022108430} \cite{CHEN2023109919} .

Recognizing that real-world scheduling is not a single-objective problem, a significant body of research has shifted towards multi-objective optimization. Early work already highlighted that nurse rostering involves balancing competing goals like cost, coverage, and staff preferences \cite{burke2004}. Modern approaches formalize this using Multi-Objective Evolutionary Algorithms (MOEAs), which generate a set of optimal trade-off solutions known as the Pareto front. The Non-dominated Sorting Genetic Algorithm II (NSGA-II) \cite{Deb2002} stands out as a state-of-the-art MOEA, frequently applied to scheduling problems for its efficiency in exploring the trade-off surface between objectives like minimizing costs and maximizing staff satisfaction.

The research frontier in MOEAs continues to advance to address increasing complexity. This includes developing strategies for large-scale multi-objective optimization with many decision variables \cite{Tian2021}, creating novel algorithms with superior performance on complex problems \cite{Sharifi2021}, and designing robust algorithms that can handle the noise and uncertainty present in real-world data \cite{Jiang2025}. Further, comparative studies emphasize that the choice of MOEA involves its own trade-off, often between computational speed and the quality of the resulting Pareto front \cite{RAHIMI2023110472}, while reviews of software frameworks highlight the need for flexible and accessible tools to advance both research and practical application \cite{pr12050869}.

Despite the sophistication of existing academic models, a significant gap persists between theoretical optimization and practical application in hospital management. Many published models  \cite{BURKE201471}\cite{Curtois2014} are validated on simplified or generalized instances that do not fully capture the granular, real-world constraints faced by nurse managers, such as managing diverse staff categories (RNs, LPNs, CNAs), seniority levels, and modular, appointment-driven hourly demands. This work aims to bridge that gap. We develop and apply a multi-objective genetic algorithm specifically tailored to this complex, real-world environment. Our primary contribution is the demonstration of a practical decision-support system that targets the granularity of hourly demands and navigates the trade-offs between three critical objectives: Operational Efficiency, Quality of Patient Care, and Staff Satisfaction. By producing a set of high-quality, non-dominated schedule options (the Pareto front), our model empowers hospital administrators to make informed, data-driven decisions that balance competing priorities, translating an advanced optimization technique into an actionable management tool.

Remaining paper is structured as follows: Section 2 provides a formal mathematical formulation of the multi-objective problem. Section 3 details the design of our Genetic Algorithm. Section 4 presents the experimental setup and discusses the results. Finally, Section 5 concludes the paper with some directions for future work.



\section{Problem Formulation}
In this paper, we address the Workforce Scheduling Problem (WSP), which involves assigning a set of employees to a series of time slots over a defined scheduling horizon. The primary challenge is to meet the fluctuating demand while respecting a complex set of operational and contractual rules. We formulate this as a multi‐objective optimization task, aiming to find a set of optimal trade‐off schedules that balance competing business goals.

\subsection{Mathematical Model Definition}
We define the schedule over a set of discrete time slots. The planning horizon consists of multiple days, where each day is divided into 48 slots of 30‐minute intervals each. The key entities are:
\begin{itemize}
  \item \textbf{Employees ($E$):} a set of employees, each with specific attributes such as job role and contractual limitations.
  \item \textbf{Days ($D$):} a set of consecutive days in the scheduling period.
  \item \textbf{Time Slots ($T$):} the set of 48 slots in each day, $T = \{0,1,\dots,47\}$.
  \item \textbf{Demand ($R_{dt}$):} staffing requirements for each day $d\in D$ and time slot $t\in T$, comprising minimum ($R_{dt}^{\min}$) and ideal ($R_{dt}^{\mathrm{ideal}}$) levels.
\end{itemize}

Our decision variable is
\[
x_{edt} = 
\begin{cases}
1, & \text{if employee }e\in E\text{ works on day }d\in D \text{ at slot }t\in T,\\
0, & \text{otherwise.}
\end{cases}
\]

\subsection{Constraints}

A feasible schedule must satisfy the following constraints:
\begin{itemize}
    \item\textbf{Unavailability:} An employee cannot be scheduled during a time window for which they have declared unavailability.
    \begin{equation}
  \sum_{d\in D}\sum_{t\in T} x_{edt}\,U_{edt} = 0,
  \quad \forall\,e\in E.
\end{equation}
  \item \textbf{Single Continuous Shift:} An employee's daily work must consist of a single, contiguous block of time. Split shifts (e.g., working in the morning and again in the evening) are forbidden. For each $e,d$, let $T_{ed}=\{\,t\mid x_{edt}=1\}$.  If $T_{ed}\neq\emptyset$, then
    \begin{equation}
  \max\bigl(T_{ed}\bigr)
  - \min\bigl(T_{ed}\bigr)
  + 1
  = \lvert T_{ed}\rvert,
  \quad \forall\,e\in E,\;d\in D.
\end{equation}
  \item \textbf{Shift Length:}  The duration of any single work shift must be within a specified range (e.g., a minimum of 4 hours and a maximum of 9 hours).
  Any continuous block $B$ of slots for $(e,d)$ should satisfy
    \begin{equation}
  L_{\min} \le 0.5\,\lvert B\rvert \le L_{\max}
\end{equation}

  \item \textbf{Minimum Rest Between Shifts:} A mandatory minimum rest period (e.g., 8 hours) must be enforced between the end of an employee's shift on one day and the start of their shift on the following day.  
  If $T_{ed}$ and $T_{e,d+1}$ are nonempty, then
    \begin{equation}
  (24 \cdot 60)
  -\bigl(\max(T_{ed})\cdot30\bigr)
  +\bigl(\min(T_{e,d+1})\cdot30\bigr)
  \;\ge\;H_{\mathrm{rest}}^{\min}\cdot60
\end{equation}
  \item \textbf{Daily and Weekly Hour Limits:}Employees cannot exceed a maximum number of scheduled hours per day or per week.
\begin{equation}
  \sum_{t\in T}0.5\,x_{edt}
  \;\le\;H_{\mathrm{day}}^{\max},
  \quad \forall\,e\in E,\;d\in D
\end{equation}

\begin{equation}
  \sum_{d\in D}\sum_{t\in T}0.5\,x_{edt}
  \;\le\;H_{\mathrm{week}}^{\max},
  \quad \forall\,e\in E
\end{equation}

  \item \textbf{Management Coverage:} During any time slot where staff are present, at least one employee with a managerial job title must be on duty.
    \begin{equation}
  \sum_{e\in E}x_{edt}>0
  \;\implies\;
  \sum_{e\in E_{\mathrm{mgr}}}x_{edt}\ge1,
  \quad \forall\,d\in D,\;t\in T
\end{equation}
\end{itemize}

\subsection{Objective Functions}
Our goal is to find schedules that are not only feasible but also optimal across several competing business dimensions. We formulate this as a multi-objective optimization problem with three primary objective functions all of which are to be minimized. Each objective is an aggregation of related penalty scores, where individual penalties may be weighted to signify their relative importance within that category.\\
We seek to minimize the vector of objectives:
\begin{equation}
  F \;=\;\bigl(f_1,\,f_2,\,f_3\bigr)    
\end{equation}

where:
\begin{enumerate}
  \item \textbf{Minimize Staffing Cost ($f_1$):}\\
    This objective quantifies the cost of inefficient staffing. Aggregates penalties for,\\
\textit{Over-coverage:} Assigning more employees to a time slot than the ideal workload requires.\\
\textit{Zero-demand Staffing:} Assigning employees to time slots where there is no patient demand.
  \item \textbf{Minimize Service Failure ($f_2$):}\\
  This objective measures the negative impact on patient care. Aggregates penalties for,\\
\textit{Coverage Shortfall:} Failing to meet the minimum required number of staff of given skill in a given time slot.\\
\textit{Missing Manager}: Failing to have a manager/supervisor on duty during an operational time slot.
  \item \textbf{Minimize Employee Dissatisfaction ($f_3$):}\\
    Aggregates soft‐constraint violations (unavailability, split shifts, undesirable lengths, rest violations, and hour‐limit exceedance). 

More details on the formulation can be found in Appendix A. 
\end{enumerate}

\section{Methodology: The Multi‐Objective GA}
\label{sec:methodology}

To solve this multi‐objective WSP, we propose a methodology based on the Non‐dominated Sorting Genetic Algorithm II (NSGA‐II), a widely recognized and effective evolutionary algorithm for multi‐objective optimization problems. Unlike single‐objective GAs that find a single best solution by optimizing a weighted‐sum fitness function, NSGA‐II evolves a population of solutions to find a set of optimal trade‐offs, known as the Pareto front. On this front, no solution can be improved in one objective without degrading performance in at least one other objective. This provides decision‐makers with a range of high‐quality, diverse schedules to choose from. A general overview of the solution approach is shown in Table~\ref{tab:nsga2-pseudocode}.


\subsection{Population Initialization}
A single solution, or \emph{chromosome}, in the population represents a complete workforce schedule for the entire planning horizon. We encode this as a dictionary where each key is a tuple \((\text{store\_id}, \text{date})\). The corresponding value is a list of 48 lists, representing the 30‐minute time slots for that day. Each of these 48 inner lists contains the unique IDs of the employees assigned to work during that specific time slot. To ensure a robust starting point for the evolutionary process, we employ a hybrid initialization strategy. We seed the initial population with 50\% of individuals generated by a greedy heuristic and 50\% generated randomly.  
The greedy heuristic attempts to build sensible schedules by assigning available employees to time slots with unmet demand.  
The random generation creates diverse schedules by assigning random shift lengths and start times to employees, ensuring a broad exploration of the solution space.

\subsection{Fitness Evaluation}
We evaluate the fitness of each chromosome not as a single scalar value, but as a fitness vector
\begin{equation}
\mathbf{F}(i) = \bigl(f_1(i),\,f_2(i),\,f_3(i)\bigr),    
\end{equation}

corresponding to the three objective functions defined in Appendix A. For each schedule in the population, we calculate the 10 individual penalty scores. These scores are then aggregated to produce the three objective values: \(f_1\) (Cost), \(f_2\) (Service Failure), and \(f_3\) (Employee Dissatisfaction). This vector is used to evaluate the dominance and quality of each solution without combining them into a single weighted sum.

\subsection{Genetic Operators}
We drive the evolution from one generation to the next using selection, crossover, and mutation operators. we have customized the operators adapted from canonical genetic algorithms to suit our specific schedule representation.

\subsubsection{Selection using Tournament Selection}
Our selection mechanism is based on the core principles of NSGA‐II. We use a binary tournament where two solutions are randomly chosen from the population. The winner is selected based on the following criteria:
\begin{itemize}
  \item If one solution has a better non‐domination rank than the other, it is chosen.
  \item If both solutions have the same rank, the one with the greater crowding distance is chosen to promote diversity on the Pareto front.
\end{itemize}

\subsubsection{Crossover Operators}
To create offspring, we select two parent solutions and combine them using one of several crossover operators, chosen probabilistically:
\begin{itemize}
  \item \textbf{Day‐Point Crossover:} A random day is selected as a crossover point. The child schedule is created by taking all days before the point from the first parent and all days after the point from the second parent.
  \item \textbf{Uniform Crossover:} For each time slot in the schedule, a random decision is made whether to inherit the set of assigned employees from the first or the second parent.
  \item \textbf{Two‐Point Slot Crossover:} The entire schedule is flattened into a single array of time slots. Two random points are chosen, and the segment of the schedule between these points is swapped between the parents.
\end{itemize}

\subsubsection{Mutation Operators}
After crossover, we apply mutation to the offspring to introduce new genetic material. One of the following mutation operators is applied with a small probability:
\begin{itemize}
  \item \textbf{Swap Employee:} In a random time slot, one assigned employee is swapped with another available employee.
  \item \textbf{Move Shift:} An entire contiguous work block for an employee is moved slightly earlier or later in the day.
  \item \textbf{Change Shift Length:} A work block is randomly made slightly longer or shorter, within the allowed shift length constraints.
\end{itemize}

\subsection{Elitism and Generation Transition}
Our methodology uses a sophisticated elitist strategy. We combine the current parent population and the generated offspring population. This combined population is then sorted into non‐domination fronts. The new parent population is formed by selecting all solutions from the best front (Front 1), then all solutions from the next best front (Front 2), and so on, until the population size is reached. This ensures that the best‐found solutions are always preserved and carried over to the next generation.
\begin{table}[ht]
  \centering
  \caption{Pseudocode of NSGA-II for Workforce Scheduling}
  \label{tab:nsga2-pseudocode}
  \begin{tabular}{p{0.05\linewidth} p{0.9\linewidth}}
    \toprule
    \textbf{Step} & \textbf{Operation} \\
    \midrule
    1 &
      \textbf{Input:}
      Population Size ($N$), Max Generations ($G_{\max}$), Problem Parameters ($P$). \\
    2 &
      \textbf{Initialization:}
      $P_0 \leftarrow \text{InitializePopulation}(N,P)$; \\
      & \quad EvaluateFitness($P_0$); \\
      & \quad $(\mathcal{F}_1,\mathcal{F}_2,\dots)\leftarrow\text{FastNonDominatedSort}(P_0)$; \\
      & \quad CalculateCrowdingDistance on each front; \\
      & \quad $t\leftarrow0$. \\
    3 &
      \textbf{While} $t < G_{\max}$: \\
      & \quad $Q_t\leftarrow\text{CreateOffspringPopulation}(P_t)$; \\
      & \quad EvaluateFitness($Q_t$); \\
      & \quad $R_t\leftarrow P_t\cup Q_t$; \\
      & \quad $(\mathcal{F}_1,\mathcal{F}_2,\dots)\leftarrow\text{FastNonDominatedSort}(R_t)$; \\
      & \quad $P_{t+1}\leftarrow\emptyset$, $i\leftarrow1$. \\
    4 &
      \textbf{Repeat} while $|P_{t+1}| + |\mathcal{F}_i| \le N$: \\
      & \quad CalculateCrowdingDistance($\mathcal{F}_i$); \\
      & \quad $P_{t+1}\leftarrow P_{t+1}\cup\mathcal{F}_i$; \\
      & \quad $i\leftarrow i+1$. \\
    5 &
      \textbf{Sort} the remaining front $\mathcal{F}_i$ by crowding distance; \\
      & \quad Fill $P_{t+1}$ up to $N$ individuals from $\mathcal{F}_i$; \\
      & \quad $t\leftarrow t+1$. \\
    \midrule
    \textbf{Output} &  Return the final Pareto front $\mathcal{F}_1$ from $P_{G_{\max}}$. \\
    \bottomrule
  \end{tabular}
\end{table}





 
\section{Results and Discussion}


While the proposed methodology has been successfully implemented in real-world hospital and retail environments, the results presented here are derived from a synthetic dataset to honor data confidentiality agreements. This dataset, which closely models these operational realities, was used to test our Genetic Algorithm (GA) across five simulated hospital units with varying sizes and requirements.

The synthetic data and case study are based on the following assumptions:
\begin{itemize}
\item The scheduling horizon is one week for all units.
\item Staffing requirements are dynamic, with typical operating hours from 6:00 AM to 11:30 PM and peak demand occurring during midday and on weekends.
\item Staff are categorized into four skill levels, denoted L1 through L4.
\item For each 30-minute interval, the input specifies both the optimal and the absolute minimum number of staff required for each skill level.
\item The problem instance incorporates employee-specific constraints, including pre-approved days off, shift preferences (see unavailable hours in  Figure~\ref{fig:schedule}), and mandatory skill coverage requirements on the unit floor (refer to Table~\ref{tab:unit_inputs} in the appendix for total number of employees in each category across units).
\end{itemize}

The configuration parameters for the genetic algorithms, as defined in our technical configuration file, are detailed in Table ~\ref{tab:ga_config}.

\begin{table}[ht]
\centering
\caption{Genetic Algorithm Configuration Parameters}
\label{tab:ga_config}
\begin{tabular}{ll}
\hline
\textbf{Parameter} & \textbf{Value} \\
\hline
Population Size & 400 \\
Number of Generations & 50 \\
Elitism Count (for Single-GA) & 2 \\
Selection Type & Tournament Selection \\
Tournament Size & 2 \\
Crossover Type & Two-Point Slot \\
Mutation Rate & 0.1 \\
\hline
\end{tabular}
\end{table}

For brevity, this section first details the results for a single representative unit to illustrate the solution's features. It then compares the performance of our proposed method against a greedy algorithm, which approximates manual scheduling practices, across multiple units before discussing the overall findings.

\subsection{Illustrative results for unit 1: Balanced Objectives}

\begin{figure}[ht]
\centering
\includegraphics[width=9cm]{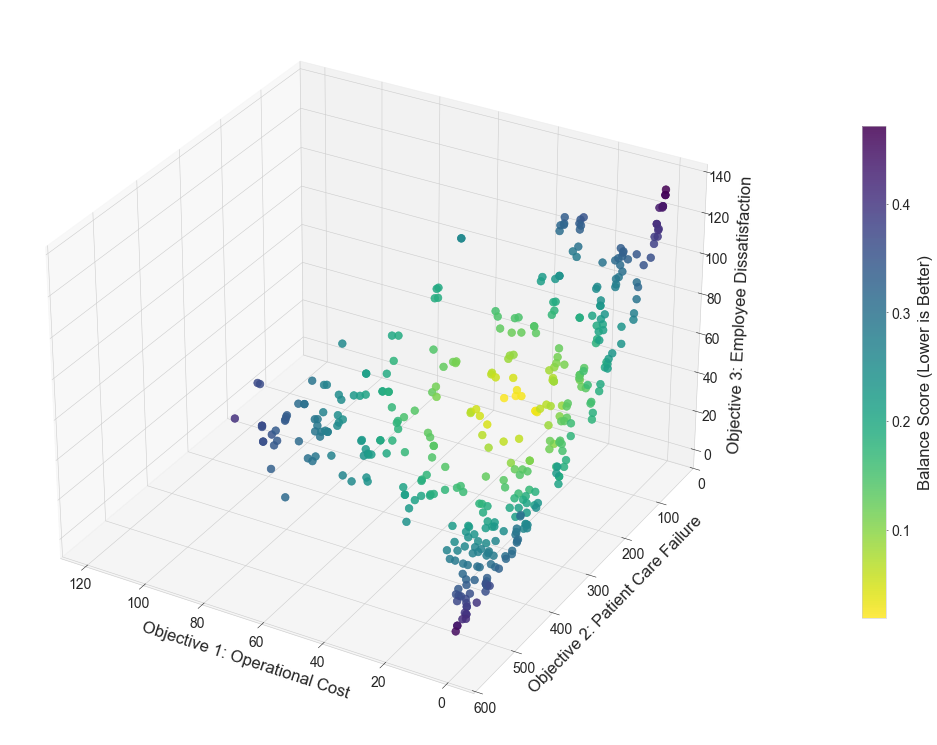}
\caption{Pareto front of the three-objective hospital staffing optimization. The 3D scatter plot shows trade-offs among (i) operational cost (penalty for overstaffing), (ii) patient care failure (penalty for understaffing relative to demand), and (iii) staff dissatisfaction (penalty for violating leave requests and preferences). Each point is a Pareto-optimal solution; yellow points highlight those achieving a balanced compromise across all three objectives. A manager may select any point on the front to best match operational priorities and quality-of-care requirements.}
\label{fig:pareto}
\end{figure}

Figure \ref{fig:pareto} shows the Pareto front of non-dominated solutions obtained by the algorithm. To showcase the solver's effectiveness, a single representative solution is selected from this front for detailed analysis. While all points on the Pareto front are equally optimal, a practical choice is the one offering the best compromise across all objectives. We define this as the "balanced" solution—the point that minimizes extreme trade-offs. For our analysis, we selected the solution with the minimum standard deviation across its normalized objective values.

\begin{figure*}[ht]
\centering
\includegraphics[width=16cm]{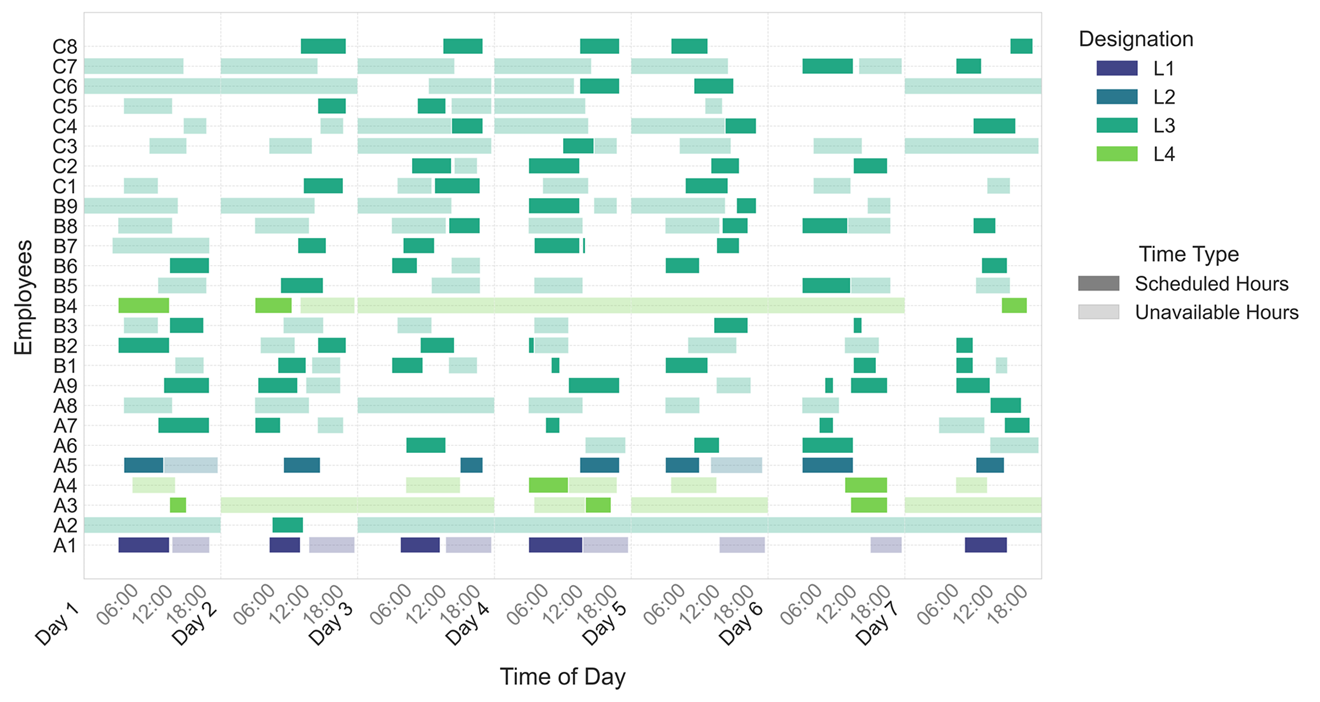}
\caption{Staff scheduling Gantt chart for unit 1. The timeline displays, for each staff member (y-axis), their assigned shifts and unavailability over the planning horizon (x-axis, in half-hour increments). Blocks are color-coded by clinical role L1,L2,L3,L4, with darker, opaque bars indicating scheduled shift periods and lighter, semi-transparent bars marking times when the employee is unavailable due to leave or other constraints.
}
\label{fig:schedule}
\end{figure*}

\begin{figure}[ht]
\centering
\includegraphics[width=9cm]{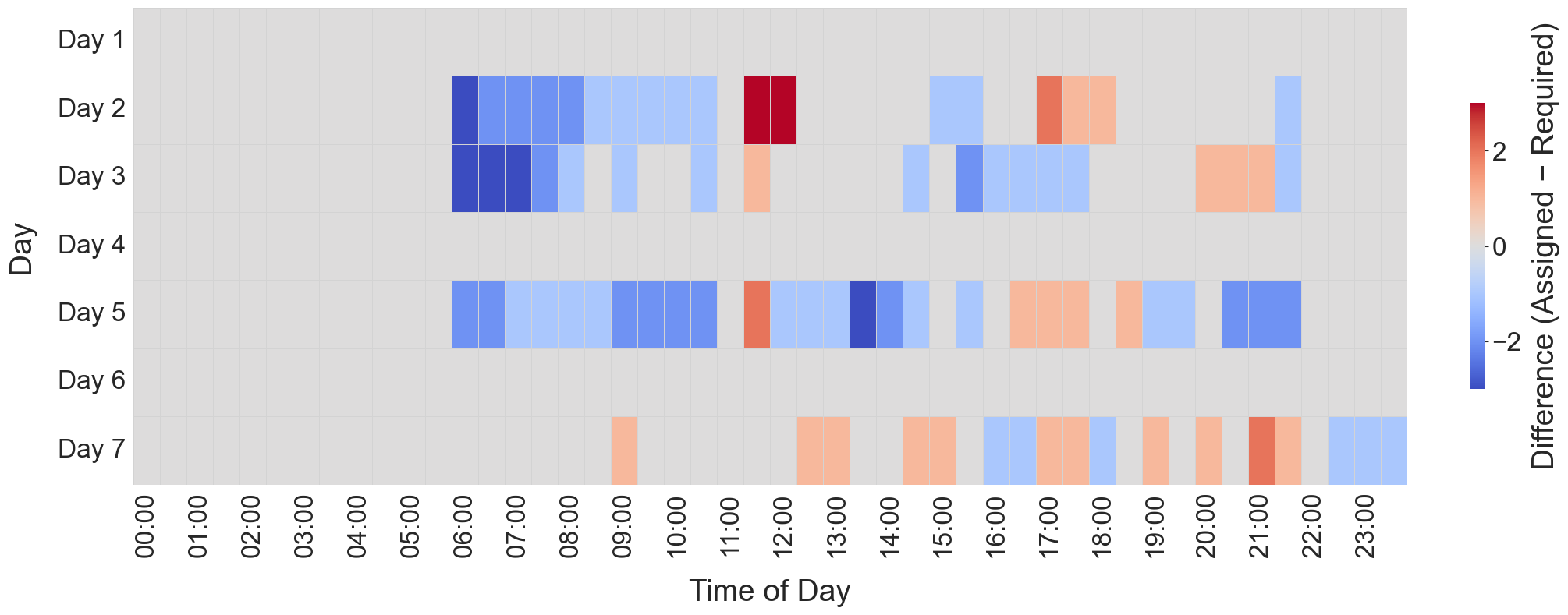}
\caption{Weekly staffing coverage heatmap for unit 1. The matrix displays, for each half-hour time slot (x-axis) over a seven-day horizon (y-axis), the difference between the number of staff assigned and the number required. Warm colors (red) denote positive values (overstaffing), while cool colors (blue) denote negative values (understaffing). The intensity of each cell corresponds to the magnitude of the deviation from demand.
}
\label{fig:Coverage}
\end{figure}

\begin{figure}[ht]
\centering
\includegraphics[width=9cm]{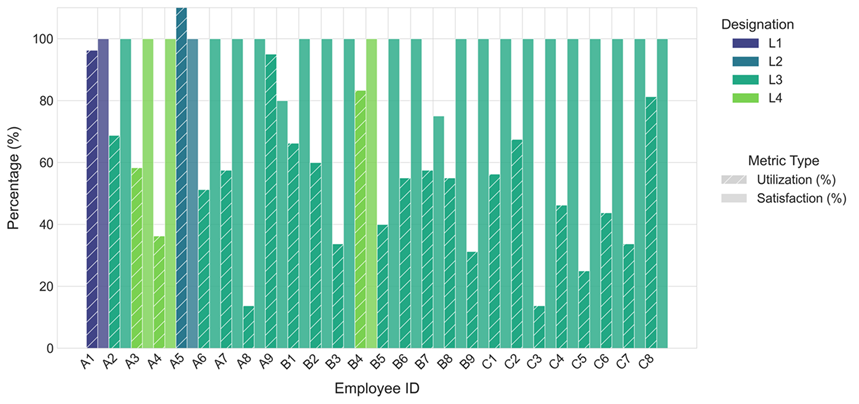}
\caption{Staff utilization and satisfaction percentages for unit 1. A grouped bar chart in which, for each employee (x-axis), two adjacent bars display (i) the utilization percentage of their contracted hours and (ii) the satisfaction percentage based on honored shift‐and‐leave preferences. Distinct bar patterns differentiate utilization from satisfaction, facilitating side-by-side comparison of operational efficiency and employee contentment.}
\label{fig:utilization}
\end{figure}

The following observations are based on a detailed analysis of this balanced solution, as presented in Figures \ref{fig:schedule},\ref{fig:Coverage}, and \ref{fig:utilization}.

\begin{itemize}
\item Patient Care:
As shown in Figure \ref{fig:Coverage}, the required staffing level is met perfectly on days 1, 4, and 6. Minor fluctuations on other days are a result of balancing staff satisfaction and operational costs. A slightly higher level of understaffing occurs for a few hours at the start of days 2 and 3, which corresponds to multiple staff members' requests not to work during those times. Nevertheless, the balanced solution effectively addresses patient needs, achieving an overall required coverage rate of 98\%.

\item Operational Cost:
Figure \ref{fig:Coverage} also illustrates that overstaffing is minimized, which directly helps to optimize operational costs. Furthermore, as shown in Figure \ref{fig:utilization}, the utilization percentage for most staff members exceeds 40\%. This is a favorable outcome, considering the stringent constraints on employee assignments and the objective of not exceeding shift requirements. This indicates the algorithm's ability to effectively manage assignments and utilization according to demand. While staff utilization could be further enhanced by reducing the total number of staff, such a strategy would risk jeopardizing patient care during periods of high demand.

\item Staff Satisfaction:
The schedule in Figure \ref{fig:schedule} confirms that staff unavailability requests are respected in most cases. The prevalence of unbroken shifts on any given day promotes operational stability. Care is also taken to maintain consistent daily working hours for staff wherever possible. This contributes to an overall staff satisfaction score of nearly 98\%, as seen in Figure \ref{fig:utilization}.
\end{itemize}

Overall, the balanced solution provides a robust schedule that managers can readily implement. For comparison, an alternative solution prioritizing operational cost is provided in the appendix B. The next section will demonstrate the superiority of the GA approach over a greedy algorithm, which simulates typical manual scheduling practices.

\subsection{Practical Improvement over manual scheduling}
To evaluate our algorithm's effectiveness, we compared its performance against a baseline greedy algorithm and a Single-Objective Genetic Algorithm (SOGA) across five distinct hospital units. The greedy algorithm was designed to emulate the typical manual scheduling process in such environments. The comparative experiment was structured as follows:
\begin{itemize}
\item The Greedy algorithm was executed 1000 times to establish a robust performance baseline.
\item The Single-Objective GA was executed 5 times to account for its stochastic nature.
\item Our Multi-Objective GA (MOO-GA) was executed 5 times to generate sets of Pareto-optimal fronts.
\end{itemize}

The results, summarized in Table~\ref{tab:unit_pivot}, reveal a clear performance hierarchy. The SOGA achieves a 61\% improvement in total fitness score over the greedy baseline. More notably, the "best balanced" solution selected from our MOO-GA's Pareto front outperforms the best SOGA result by an additional 18\%.

This finding demonstrates a key advantage of our approach. The multi-objective search not only identifies effective trade-off solutions but also discovers schedules that are globally superior, even when assessed by a single, aggregated metric. This confirms that the MOO-GA offers significant advantages over both the SOGA and the baseline greedy algorithm.

\begin{table}[ht]
  \centering
  \small
  \setlength{\tabcolsep}{4pt}
  \caption{Unit‐wise comparison of Greedy, SOGA, and MOO Balanced outcomes}
  \label{tab:unit_pivot}
  \begin{tabularx}{\columnwidth}{@{} l l *{4}{c} @{}}
    \toprule
    \textbf{Unit ID} & \textbf{Method}
      & \textbf{Total Fitness}
      & \textbf{Obj $f_1$}
      & \textbf{Obj $f_2$}
      & \textbf{Obj $f_3$} \\
    \midrule
    \multirow{3}{*}{Unit 1}
      & Greedy           & 796.32  &   0   & 710.84 &  85.47 \\
      & SOGA    & 312.5   & 60.6  & 200    &  51.9  \\
      & MOO Balanced     & 255.8   & 49.2  & 169    &  37.6  \\
    \addlinespace
    \multirow{3}{*}{Unit 2}
      & Greedy           & 474.67  &   0   & 440.37 &  34.31 \\
      & SOGA    & 162.3   & 31.2  & 75     &  56.1  \\
      & MOO Balanced     & 161.3   & 37.2  & 100.4  &  23.7  \\
    \addlinespace
    \multirow{3}{*}{Unit 3}
      & Greedy           & 753.6   &   0   & 671.59 &   82   \\
      & SOGA    & 297     & 54    & 151.4  &  91.6  \\
      & MOO Balanced     & 283.3   & 74    & 151.2  &  58.1  \\
    \addlinespace
    \multirow{3}{*}{Unit 4}
      & Greedy           & 457.32  &   0   & 439.25 &  18.08 \\
      & SOGA    & 142.5   & 27    & 85     &  30.5  \\
      & MOO Balanced     & 124.4   & 18    & 86.2   &  20.2  \\
    \addlinespace
    \multirow{3}{*}{Unit 5}
      & Greedy           & 1094.49 &   0   & 994.5  &  99.99 \\
      & SOGA    & 451.8   & 72.4  & 262    & 117.4  \\
      & MOO Balanced     & 432.8   & 96.4  & 263.8  &  72.6  \\
    \bottomrule
    \end{tabularx}
\end{table}

The results confirm the significant practical utility of the proposed Multi-Objective Genetic Algorithm (MOO-GA) for complex scheduling problems, particularly within the demanding healthcare environment. By outperforming both greedy and single-objective methods, our approach demonstrates its ability to navigate the intricate trade-offs between minimizing operational costs, maximizing patient care coverage, and ensuring staff satisfaction. The algorithm provides hospital administrators not with a single, rigid schedule, but with a Pareto-optimal set of high-quality solutions. This empowers them to make informed, data-driven decisions that can be adapted to fluctuating priorities—for instance, prioritizing patient coverage during a surge or focusing on cost containment during a quieter period.

The fundamental principles of this model extend well beyond healthcare. Many industries, such as retail and warehouse logistics, face analogous multi-objective scheduling challenges. In retail, the goal is to align staffing levels with dynamic customer footfall to optimize service without overspending on labor. Similarly, warehouse operations must schedule staff to meet fluctuating order volumes and fulfillment deadlines. In both sectors, accommodating employee preferences is increasingly vital for retention and morale. The demonstrated success of our GA in balancing these competing objectives suggests that this framework can be readily adapted to these environments, offering a robust and flexible tool for efficient workforce management in any service-driven industry characterized by dynamic demand and diverse operational constraints.

\section{Conclusion}
In this paper, we addressed the complex, multi-objective challenge of workforce scheduling within the healthcare sector. We developed and implemented a Multi-Objective Genetic Algorithm (MOO-GA) designed to simultaneously optimize for operational cost, patient care coverage, and staff satisfaction. The model was validated using a comprehensive, synthetic dataset meticulously crafted to reflect the real-world operational dynamics of a hospital, including fluctuating demand, diverse staff skill sets, and specific employee constraints.

The results clearly demonstrate the superiority of our multi-objective approach. When compared against both a traditional greedy algorithm and a Single-Objective GA, our MOO-GA consistently identified more effective and balanced scheduling solutions. Notably, it produced a set of Pareto-optimal schedules, offering administrators a range of viable options rather than a single, rigid output. This provides a crucial decision-support framework, enabling managers to flexibly adapt to changing priorities. The findings confirm that this methodology is not only a practical tool for improving operational efficiency and quality of care in hospitals but also holds significant potential for application in other service-driven industries, such as retail and logistics, that face similar dynamic staffing challenges.



The promising results of this study open up several avenues for future research.
\begin{itemize}
    \item \textbf{Hybridization with Local Search:} To improve the fine-tuning of solutions, the proposed GA could be hybridized with local search techniques. A Memetic Algorithm, for instance, could apply a local search heuristic to the solutions in each generation, potentially accelerating convergence towards higher-quality regions of the Pareto front.
\item \textbf{Dynamic and Real-Time Scheduling:} The current model operates on a static set of inputs. A significant extension would be to adapt the framework for dynamic scheduling to handle real-time disruptions, such as employee absences or unexpected surges in demand, by allowing for rapid re-optimization of the existing schedule. Integration of advanced techniques such as graphs neural networks and Reinforcement Learning is found promising and needs to be explored. 
\item \textbf{Integration of Decision-Maker Preferences:} While our method for selecting a "balanced" solution is effective for analysis, future work could explore incorporating decision-maker preferences directly into the evolutionary process. Techniques like Reference Point-based NSGA-II could guide the search towards a specific region of the Pareto front that is of most interest to the user.
\item \textbf{Integrate Machine Learning for improved capability} Integrating machine learning models to forecast key inputs, such as customer demand and employee availability patterns, will provide the optimization algorithm with more accurate and probabilistic data to work with.
\end{itemize}






\begin{thebibliography}{24}
\providecommand{\natexlab}[1]{#1}
\providecommand{\url}[1]{\texttt{#1}}
\expandafter\ifx\csname urlstyle\endcsname\relax
  \providecommand{\doi}[1]{doi: #1}\else
  \providecommand{\doi}{doi: \begingroup \urlstyle{rm}\Url}\fi

\bibitem[Azaiez and {Al Sharif}(2005)]{AZAIEZ2005491}
M.~Azaiez and S.~{Al Sharif}.
\newblock A 0-1 goal programming model for nurse scheduling.
\newblock \emph{Computers \& Operations Research}, 2005.

\bibitem[Burke et~al.(2004)Burke, De~Causmaecker, and Vanden~Berghe]{burke2004}
E.~Burke, P.~De~Causmaecker, and G.~Vanden~Berghe.
\newblock Novel metaheuristic approaches to nurse rostering problems in belgian hospitals.
\newblock \emph{Handbook of Scheduling: Algorithms, Models, and Performance Analysis}, 2004.

\bibitem[Burke and Curtois(2014)]{BURKE201471}
E.~K. Burke and T.~Curtois.
\newblock New approaches to nurse rostering benchmark instances.
\newblock \emph{European Journal of Operational Research}, 2014.

\bibitem[Curtois and Qu(2014)]{Curtois2014}
T.~Curtois and R.~Qu.
\newblock Computational results on new staff scheduling benchmark instances.
\newblock Technical report, ASAP Research Group, School of Computer Science, University of Nottingham, NG8 1BB, Nottingham, UK, 2014.

\bibitem[Deb et~al.(2002)Deb, Pratap, Agarwal, and Meyarivan]{Deb2002}
K.~Deb, A.~Pratap, S.~Agarwal, and T.~Meyarivan.
\newblock A fast and elitist multiobjective genetic algorithm: Nsga-ii.
\newblock \emph{IEEE Transactions on Evolutionary Computation}, 2002.

\bibitem[Ernst et~al.(2004)Ernst, Jiang, Krishnamoorthy, and Sier]{ERNST2004}
A.~Ernst, H.~Jiang, M.~Krishnamoorthy, and D.~Sier.
\newblock Staff scheduling and rostering: A review of applications, methods and models.
\newblock \emph{European Journal of Operational Research}, 2004.

\bibitem[Jiang et~al.(2025)Jiang, Gao, Zhu, and Xu]{Jiang2025}
W.~Jiang, K.~Gao, S.~Zhu, and L.~Xu.
\newblock A novel robust multi-objective evolutionary optimization algorithm based on surviving rate.
\newblock \emph{Complex \& Intelligent Systems}, 2025.

\bibitem[Kumar et~al.(2014)Kumar, Nagalakshmi, and Kumaraguru]{Kumar2014}
B.~S. Kumar, G.~Nagalakshmi, and S.~Kumaraguru.
\newblock A shift sequence for nurse scheduling using linear programming problem.
\newblock \emph{IOSR Journal of Nursing and Health Science}, 2014.

\bibitem[Liu and Zhang(2021)]{Tianyu2021}
T.~Liu and L.~Zhang.
\newblock Apply artificial neural network to solving manpower scheduling problem.
\newblock \emph{International Conference on Big Data and Artificial Intelligence (BDAI)}, 2021.

\bibitem[McKinsey and Company(2023)]{McKinsey2023}
McKinsey and Company.
\newblock Nursing in 2023: How the profession is evolving.
\newblock Technical report, McKinsey \& Company, 2023.

\bibitem[Oliveira et~al.(2024)Oliveira, Rocha, and Alves]{Oliveira2024}
M.~Oliveira, A.~M. A.~C. Rocha, and F.~Alves.
\newblock Using or-tools when solving the nurse scheduling problem.
\newblock \emph{Springer Nature Switzerland}, 2024.

\bibitem[Pahlevanzadeh et~al.(2021)Pahlevanzadeh, Jolai, Goodarzian, and Ghasemi]{Pahlevanzadeh2021}
M.~J. Pahlevanzadeh, F.~Jolai, F.~Goodarzian, and P.~Ghasemi.
\newblock A new two-stage nurse scheduling approach based on occupational justice considering assurance attendance in works shifts by using z-number method: A real case study.
\newblock \emph{RAIRO-Oper. Res.}, 2021.

\bibitem[Pătrăușanu et~al.(2024)Pătrăușanu, Florea, Neghină, Dicoiu, and Chiș]{pr12050869}
A.~Pătrăușanu, A.~Florea, M.~Neghină, A.~Dicoiu, and R.~Chiș.
\newblock A systematic review of multi-objective evolutionary algorithms optimization frameworks.
\newblock \emph{Processes}, 2024.

\bibitem[Rahimi et~al.(2023)Rahimi, Gandomi, Nikoo, and Chen]{RAHIMI2023110472}
I.~Rahimi, A.~H. Gandomi, M.~R. Nikoo, and F.~Chen.
\newblock A comparative study on evolutionary multi-objective algorithms for next release problem.
\newblock \emph{Applied Soft Computing}, 2023.

\bibitem[Rahimian et~al.(2017)Rahimian, Akartunalı, and Levine]{RAHIMIAN201783}
E.~Rahimian, K.~Akartunalı, and J.~Levine.
\newblock A hybrid integer and constraint programming approach to solve nurse rostering problems.
\newblock \emph{Computers \& Operations Research}, 2017.

\bibitem[Sharifi et~al.(2021)Sharifi, Akbarifard, Qaderi, and Madadi]{Sharifi2021}
M.~R. Sharifi, S.~Akbarifard, K.~Qaderi, and M.~R. Madadi.
\newblock A new optimization algorithm to solve multi-objective problems.
\newblock \emph{Scientific Reports}, 2021.

\bibitem[Talarico and Duque(2015)]{TALARICO201565}
L.~Talarico and P.~A.~M. Duque.
\newblock An optimization algorithm for the workforce management in a retail chain.
\newblock \emph{Computers \& Industrial Engineering}, 2015.

\bibitem[Tian et~al.(2021)Tian, Si, Zhang, Cheng, He, Tan, and Jin]{Tian2021}
Y.~Tian, L.~Si, X.~Zhang, R.~Cheng, C.~He, K.~C. Tan, and Y.~Jin.
\newblock Evolutionary large-scale multi-objective optimization: A survey.
\newblock \emph{ACM Computing Surveys}, 2021.

\bibitem[Uwe and Kathryn(2004)]{AICKELIN2004761}
A.~Uwe and A.~D. Kathryn.
\newblock An indirect genetic algorithm for a nurse-scheduling problem.
\newblock \emph{Computers \& Operations Research}, 2004.

\bibitem[Van~den Bergh et~al.(2013)Van~den Bergh, Beliën, De~Bruecker, Demeulemeester, and De~Boeck]{VANDENBERGH2013367}
J.~Van~den Bergh, J.~Beliën, P.~De~Bruecker, E.~Demeulemeester, and L.~De~Boeck.
\newblock Personnel scheduling: A literature review.
\newblock \emph{European Journal of Operational Research}, 2013.

\bibitem[Wasanapradit et~al.(2011)Wasanapradit, Mukdasanit, Chaiyaratana, and Srinophakun]{Wasanapradit2011}
T.~Wasanapradit, N.~Mukdasanit, N.~Chaiyaratana, and T.~Srinophakun.
\newblock Solving mixed-integer nonlinear programming problems using improved genetic algorithms.
\newblock \emph{Korean Journal of Chemical Engineering}, 2011.

\bibitem[Wong et~al.(2014)Wong, Xu, and Chin]{WONG201499}
T.~Wong, M.~Xu, and K.~Chin.
\newblock A two-stage heuristic approach for nurse scheduling problem: A case study in an emergency department.
\newblock \emph{Computers \& Operations Research}, 2014.

\bibitem[Ziyi et~al.(2022)Ziyi, Yajie, and Patrick]{CHEN2022108430}
C.~Ziyi, D.~Yajie, and D.~C. Patrick.
\newblock Neural networked-assisted method for the nurse rostering problem.
\newblock \emph{Computers \& Industrial Engineering}, 2022.

\bibitem[Ziyi et~al.(2023)Ziyi, Patrick, and Yajie]{CHEN2023109919}
C.~Ziyi, D.~C. Patrick, and D.~Yajie.
\newblock A combined mixed integer programming and deep neural network-assisted heuristics algorithm for the nurse rostering problem.
\newblock \emph{Applied Soft Computing}, 2023.

\end{thebibliography}

\section*{Appendix A: Mathematical Formulation}
\label{sec:appendix_math_model}

This appendix provides the detailed mathematical formulation of the Workforce Scheduling Problem (WSP) addressed in this study.

\subsection*{A.1 Sets and Indices}
\begin{itemize}
    \item $E$: The set of all employees, indexed by $e$.
    \item $D$: The set of all days in the scheduling horizon, indexed by $d$.
    \item $T$: The set of 48 time slots in a day, indexed by $t$, where $t \in \{0, 1, \dots, 47\}$.
    \item $E_{mgr} \subset E$: The subset of employees who are managers.
\end{itemize}

\subsection*{A.2 Parameters}
\begin{itemize}
    \item $R_{dt}^{\min}$: The minimum required number of employees for day $d$, time slot $t$.
    \item $R_{dt}^{\text{ideal}}$: The ideal number of employees for day $d$, time slot $t$.
    \item $U_{edt}$: A binary parameter, equal to 1 if employee $e$ is unavailable on day $d$ at time slot $t$, and 0 otherwise.
    \item $L_{\min}, L_{\max}$: The minimum and maximum allowed duration of a single work shift in hours.
    \item $H_{\text{day}}^{\max}, H_{\text{week}}^{\max}$: The maximum number of work hours allowed for an employee per day and per week, respectively.
    \item $H_{\text{rest}}^{\min}$: The minimum required rest hours between two consecutive shifts.
    \item $w_{\text{penalty}}$: A set of weights for each type of penalty (e.g., $w_{\text{over}}, w_{\text{shortfall}}$).
\end{itemize}

\subsection*{A.3 Decision Variables}
\[
x_{edt} = 
\begin{cases}
1, & \text{if employee }e\in E\text{ works on day }d\in D \text{ at slot }t\in T,\\
0, & \text{otherwise.}
\end{cases}
\]

\subsection*{A.4 Objective Functions}
The problem is formulated with three objective functions to be minimized simultaneously: 
\begin{equation}
    F = (f_1, f_2, f_3)
\end{equation}

\begin{enumerate}
\item \textbf{Minimize Staffing Cost ($f_1$):}
    \begin{equation}
\begin{split}
  f_1 
  &= \sum_{d\in D}\sum_{t\in T}
     w_{\mathrm{over}}
     \max\Bigl(0,\sum_{e\in E}x_{edt}-R_{dt}^{\mathrm{ideal}}\Bigr) \\
  &\quad
     +\,w_{\mathrm{zero}}\;
     \Bigl(\sum_{e\in E}x_{edt}\;\text{if }R_{dt}^{\min}=0\Bigr).
\end{split}
\end{equation}

    \item \textbf{Minimize Service Failure ($f_2$):}
    \begin{equation}
\begin{split}
  f_2 
  &= \sum_{d\in D}\sum_{t\in T}
     w_{\mathrm{shortfall}}
     \max\Bigl(0,\,R_{dt}^{\min} - \sum_{e\in E}x_{edt}\Bigr) \\
  &\quad
     +\,w_{\mathrm{mgr}}\,
     \Bigl(1 - \operatorname{sign}\bigl(\sum_{e\in E_{\mathrm{mgr}}}x_{edt}\bigr)\Bigr)
     \quad\text{if }\sum_{e\in E}x_{edt}>0.
\end{split}
\end{equation}

    \item \textbf{Minimize Employee Dissatisfaction ($f_3$):} This is a composite function representing the sum of all penalties related to schedule quality for employees.
    \begin{equation}
\begin{split}
  f_3
  &= \sum_{e\in E}\Bigl(
       \mathrm{Penalty}_{\mathrm{unavail}}
     + \mathrm{Penalty}_{\mathrm{split}}
     + \mathrm{Penalty}_{\mathrm{length}} \\
  &\qquad\quad
     + \mathrm{Penalty}_{\mathrm{rest}}
     + \mathrm{Penalty}_{\mathrm{hours}}
    \Bigr).
\end{split}
\end{equation}

\end{enumerate}

\subsection*{A.5 Constraints}
The following hard constraints must be satisfied for a schedule to be considered feasible. In our genetic algorithm, violations of these constraints are heavily penalized to guide the search towards feasibility.

\begin{itemize}
    \item \textbf{Unavailability:}
    \begin{equation}
  \sum_{d\in D}\sum_{t\in T} x_{edt}\,U_{edt} = 0,
  \quad \forall\,e\in E.
\end{equation}

    \item \textbf{Single Continuous Shift:} For any employee $e$ on any day $d$, if their set of worked slots $T_{ed} = \{t \mid x_{edt} = 1\}$ is not empty, then it must be a continuous block. This can be expressed as:
    \begin{equation}
  \max\bigl(T_{ed}\bigr)
  - \min\bigl(T_{ed}\bigr)
  + 1
  = \lvert T_{ed}\rvert,
  \quad \forall\,e\in E,\;d\in D.
\end{equation}
    \item \textbf{Shift Length:} For any continuous work block $B$ for employee $e$ on day $d$:
    \begin{equation}
  L_{\min} \le 0.5\,\lvert B\rvert \le L_{\max}
\end{equation}

    \item \textbf{Minimum Rest Between Shifts:} For any employee $e$ working on consecutive days $d$ and $d+1$:
    \begin{equation}
  (24 \cdot 60)
  -\bigl(\max(T_{ed})\cdot30\bigr)
  +\bigl(\min(T_{e,d+1})\cdot30\bigr)
  \;\ge\;H_{\mathrm{rest}}^{\min}\cdot60
\end{equation}

    \item \textbf{Daily and Weekly Hour Limits:}
\begin{equation}
  \sum_{t\in T}0.5\,x_{edt}
  \;\le\;H_{\mathrm{day}}^{\max},
  \quad \forall\,e\in E,\;d\in D
\end{equation}

\begin{equation}
  \sum_{d\in D}\sum_{t\in T}0.5\,x_{edt}
  \;\le\;H_{\mathrm{week}}^{\max},
  \quad \forall\,e\in E
\end{equation}

    \item \textbf{Management Coverage:}
    \begin{equation}
  \sum_{e\in E}x_{edt}>0
  \;\implies\;
  \sum_{e\in E_{\mathrm{mgr}}}x_{edt}\ge1,
  \quad \forall\,d\in D,\;t\in T
\end{equation}
\end{itemize}

\section*{Appendix B: Additional Information}
\label{sec:additional_info}


\begin{table}[!ht]
  \centering
  \caption{Hospital Unit wise Employee count}
  \label{tab:unit_inputs}
  \begin{tabular}{lcr}
    \hline
    \textbf{Parameter}  & \textbf{Employee Type and count}  & \textbf{Total count}\\ \hline
    Unit 1         & L1-1, L2-1, L3-21, L4-3     & 26  \\
    Unit 2         & L1-1, L2-1, L3-9, L4-3     & 14  \\
    Unit 3         & L1-1, L2-1, L3-16, L4-5     & 23  \\
    Unit 4         & L1-1, L2-1, L3-6, L4-2     & 10  \\
    Unit 5         & L1-1, L2-1, L3-25, L4-6     & 33 \\
\end{tabular}
\end{table}

\subsection*{B.1 Illustrative Results for unit 1: Staffing cost objective prioritized}

\begin{figure}[!ht]
\centering
\includegraphics[width=9cm]{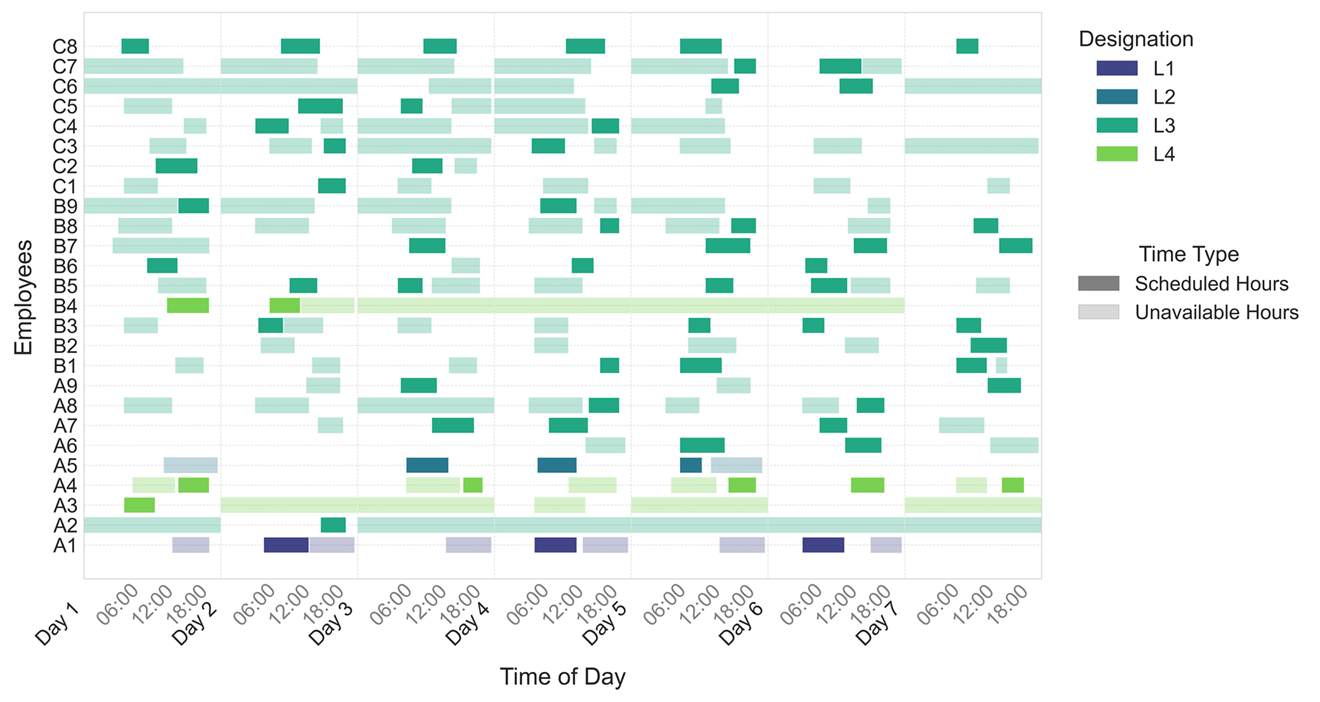}
\caption{Staff scheduling Gantt chart for unit 1 with staffing cost objective prioritized. The Assignment overall for each Nurse goes down as the solution tries to minimize the cost associated with staffing.
}
\label{fig:low_cost_schedule}
\end{figure}

\begin{figure}[!ht]
\centering
\includegraphics[width=9cm]{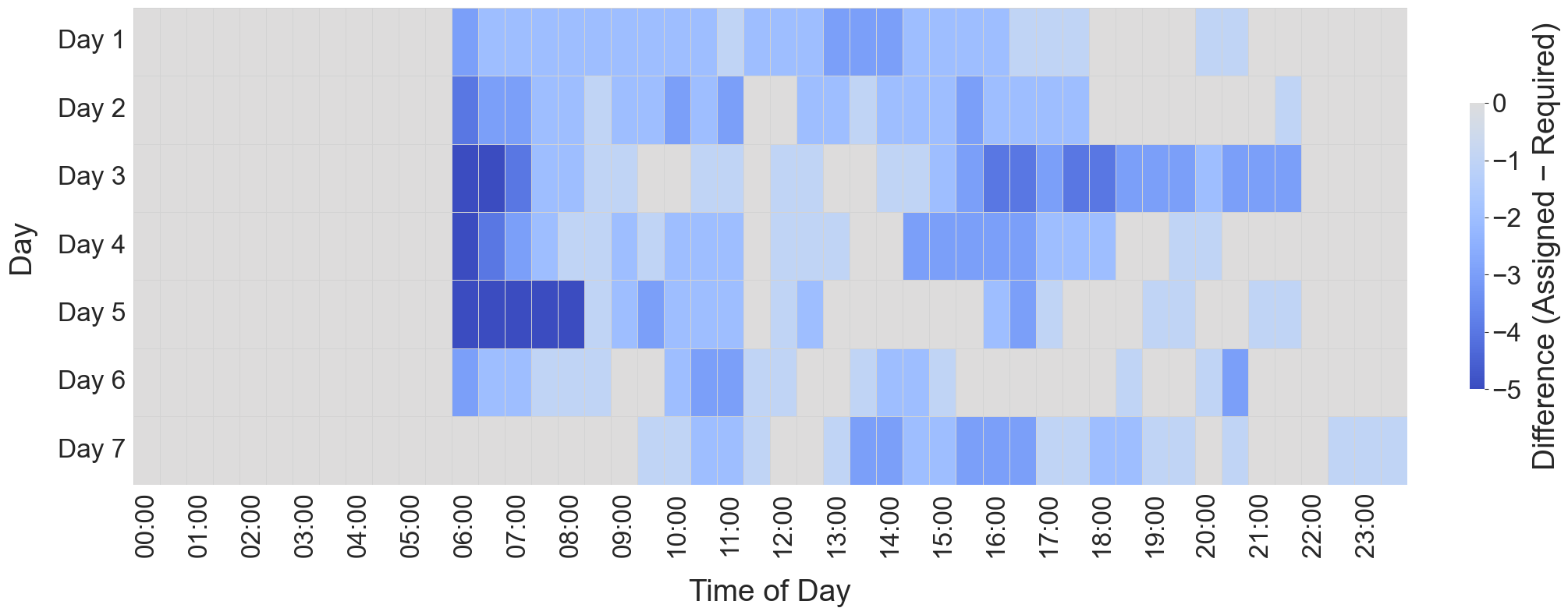}
\caption{Weekly staffing coverage heatmap for unit 1 with staffing cost objective prioritized. With Cost prioritized the Patient care goes down as the tradeoff comes with the under-assignment of nurses.
}
\label{fig:low_cost_Coverage}
\end{figure}

\begin{figure}[!ht]
\centering
\includegraphics[width=9cm]{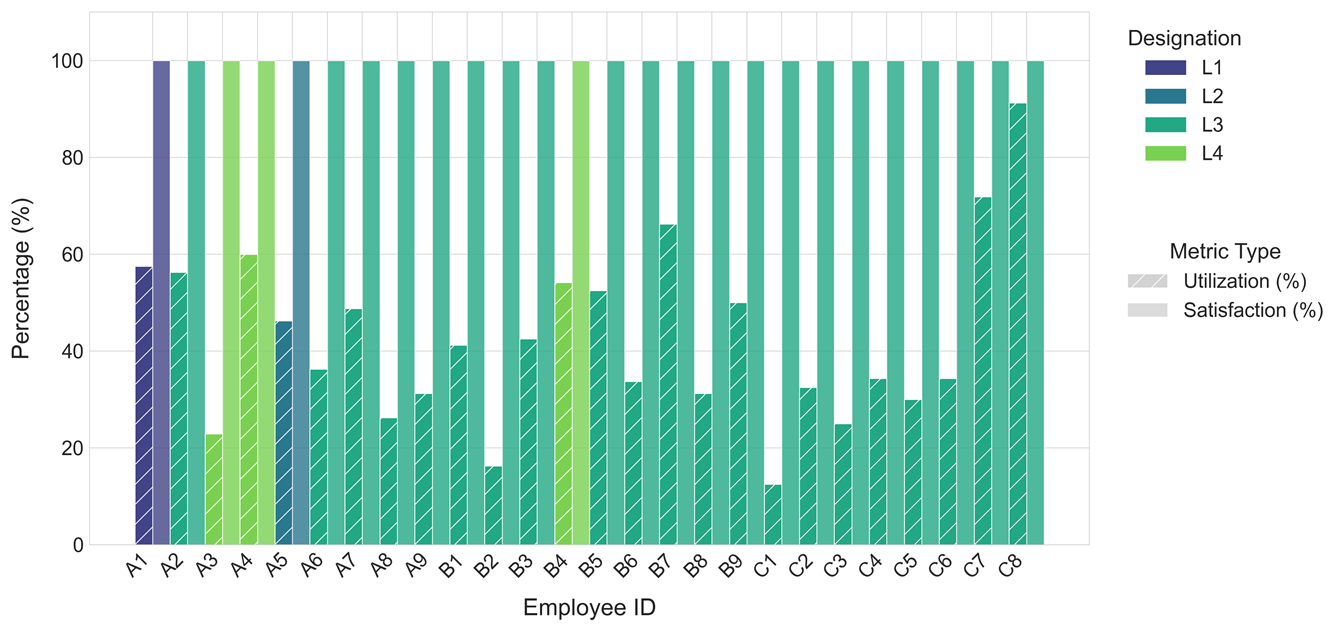}
\caption{Staff utilization and satisfaction percentages for unit 1 with staffing cost objective prioritised. Utilization can be seen going down with lower assignments and satisfaction going up as the nurses are working less.}
\label{fig:low_cost_utilization}
\end{figure}

\end{document}